# Streaming an image through the eye: The retina seen as a dithered scalable image coder

Khaled Masmoudi[a,1,*], Marc Antonini[a,1], Pierre Kornprobst[b,2]

[a]*2000 route des lucioles 06903 Sophia Antipolis, France*
[b]*2004 route des lucioles 06902 Sophia Antipolis, France*

**Abstract**

We propose the design of an original scalable image coder/decoder that is inspired from the mammalians retina. Our coder accounts for the time-dependent and also non-deterministic behavior of the actual retina. The present work brings two main contributions: As a first step, (i) we design a deterministic image coder mimicking most of the retinal processing stages and then (ii) we introduce a retinal noise in the coding process, that we model here as a dither signal, to gain interesting perceptual features. Regarding our first contribution, our main source of inspiration will be the biologically plausible model of the retina called *Virtual Retina*. The coder that we propose has two stages. The first stage is an image transform which is performed by the outer layers of the retina. Here we model it by filtering the image with a bank of difference of Gaussians with time-delays. The second stage is a time-dependent analog-to-digital conversion which is performed by the inner layers of the retina. The main novelty of this coder is to show that the time-dependent behavior of the retina cells could ensure, in an implicit way, scalability and bit allocation. Regarding our second contribution, we reconsider the inner layers of the retina. We emit a possible interpretation for the non-determinism observed by neurophysiologists in their output. For this sake, we model the retinal noise that occurs in these layers by a dither signal. The dithering process that we propose adds several interesting features to our image coder. The dither noise whitens the reconstruction error and decorrelates it from the input stimuli. Furthermore, integrating the dither noise in our coder allows a faster recognition of the fine details of the image during the decoding process. Our present paper goal is twofold. First, we aim at mimicking as closely as possible the retina for the design of a novel image coder while keeping encouraging performances. Second, we bring a new insight concerning the non-deterministic behavior of the retina.

## 1. Introduction

Research in still image compression yielded several coding algorithms as the JPEG standards [2, 5]. Though, these algorithms follow for their most a characteristic design schema. Namely, three main stages are considered. First, a transform is applied to the image. Second, the transformed data is quantized. Finally, an entropic coder compresses the bitstream to be transmitted. Interestingly, we retrieve a similar behavior in the mammalians retina [7]. Indeed, accumulated neurophysiologic evidence showed that the retina applies a transform to the image. Then, the retina binarizes the signal obtained to generate a set of uniformly shaped electrical impulses: the spikes [23]. Finally, several works tend to confirm that the retina generates a compact code to be transmitted to the visual cortex [22]. Thus, we are convinced that an interdisciplinary approach combining the signal processing techniques and the knowledge acquired by neurophysiologists would lead to novel coding algorithms beyond the standards.

---

[*]Corresponding author
[1]UNS–CNRS–I3S laboratory
[2]INRIA–NeuroMathComp project team

In order to design a novel bio-inspired image coder, we need to capture the main properties of the retina processing. Though the action of perception seems effortless, neurophysiological experiments proved that the mechanisms involved in the retina are highly complex and demanding. Recent studies such as [11] confirmed that the retina is doing non-trivial operations to the input signal before transmission to the visual cortex. Besides, the retina appears to be a non-deterministic system. The retinal code is characterized by random fluctuations. Indeed, recordings made at the output of the retina show that a single stimulus leads to different codes across trials. So that we have to deal with two issues. The first one is the complexity of the retinal processing and the need to decipher the code generated. The second issue is the non-determinism of the retinal response and its possible perceptual role. Therefore, our present paper will be structured around two parts.

In a first step, our goal is to reproduce the main stages of the retina processing for the design of a retina-inspired coder with a deterministic behavior. Our main source of inspiration will be the bio-plausible *Virtual Retina* model [36] whose goal was to find the best compromise between the biological reality and the possibility to make large-scale simulations. Based on this model, we propose a coding scheme following the architecture and functionalities of the retina. Unlike most of the efforts mimicking the retina behavior, here we focus explicitly on the coding application. So that, we do some adaptations to the model in order to be able to conceive a decoding scheme and retrieve the original stimulus. We keep our design as close as possible to biological reality taking into account the retina processing complexity, while keeping an interesting rate/distortion trade-off. The coder/decoder that we design offers several interesting features such as scalability.

In a second step, we tackle the issue of reproducing the trial-to-trial variability in the retina and its possible role. Here, we make the hypothesis that the non-determinism observed is a dithering process. So that, we identify the deep retina layers behavior to a non-subtractive dithered A/D converter. We elaborate a multiscale model for the distribution of the dither noise that we integrate in our coder. The hypothesis that we emit is seducing because of the perceptual impact it induces. Still, our model obeys the biological plausibility constraint. Interestingly, the dither noise provides our coder with perceptual properties such as the enhancement of the image contours and singularities as well as the reconstruction error whitening.

This paper is organized into two parts. The first part consists of the Sections 2 to 5 and presents the design of our bio-inspired scalable image coder/decoder with a deterministic behavior. This part is organized as follows. In Section 2 we revisit the retina model called *Virtual Retina* [36]. In Section 3, we show how this retina model can be used as the basis of a novel bio-inspired image coder. In Section 4, we present the decoding pathway. In Section 5, we show the main results that demonstrate the properties of our model. The second part consists of the Sections 6 to 7. In Section 6, we show how we integrated the dithering process in our coder/decoder. Then, in Section 7, we detail the perceptual impact of it. We show, on two test images with different properties, the ability of our dithered scalable coder to accelerate the recognition of the image details and singularities during the decoding process. Finally, in Section 8, we summarize our main conclusions.

## 2. *Virtual Retina*: A bio-plausible retina model

One first motivation for our work is to investigate the retina functional architecture and use it as a design basis to devise new codecs. So, it is essential to understand what are the main functional principles of the retina processing. The literature in computational neuroscience dealing with the retina proposes different models (see, e.g., [35] for a review). These models are very numerous, ranking from detailed models of a specific physiological phenomenon, to large-scale models of the whole retina.

In this article, we focused on the category of large-scale models as we are interested in a model that gathers the main features of the mammalians retina. Within this category, we considered the retina model called *Virtual Retina* [36]. This model is one of the most complete ones, in the literature, as it encompasses the major features of the actual mammalians retina. This model is mostly state-of-the-art and the authors confirmed its relevance by reproducing accurately actual retina cell recordings for several experiments.

The architecture of the *Virtual Retina* model follows the structure of the mammalians retina as schematized in Figure 1(a). The model has several interconnected layers and three main processing steps can be distinguished:

- The outer layers: The first processing step is described by non-separable spatio-temporal filters, behaving as time-dependent edge detectors. This is a classical step implemented in several retina models.

- The inner layers: A non-linear contrast gain control is performed. This step models mainly bipolar cells by control circuits with time-varying conductances.

- The ganglionic layer: Leaky integrate and fire neurons are implemented to model the ganglionic layer processing that finally converts the stimulus into spikes.

Given this model as a basis, our goal is to adapt it to conceive the new codec presented in the next sections.

## 3. The coding pathway

The coding pathway is schematized in Figure 1(b). It follows the same architecture as *Virtual Retina*. However, since we have to define also a decoding pathway, we need to think about the invertibility of each processing stage. For this reason some adaptations are required and described in this section. The coder design, presented in this section, is an enhancement of our previous effort in [17].

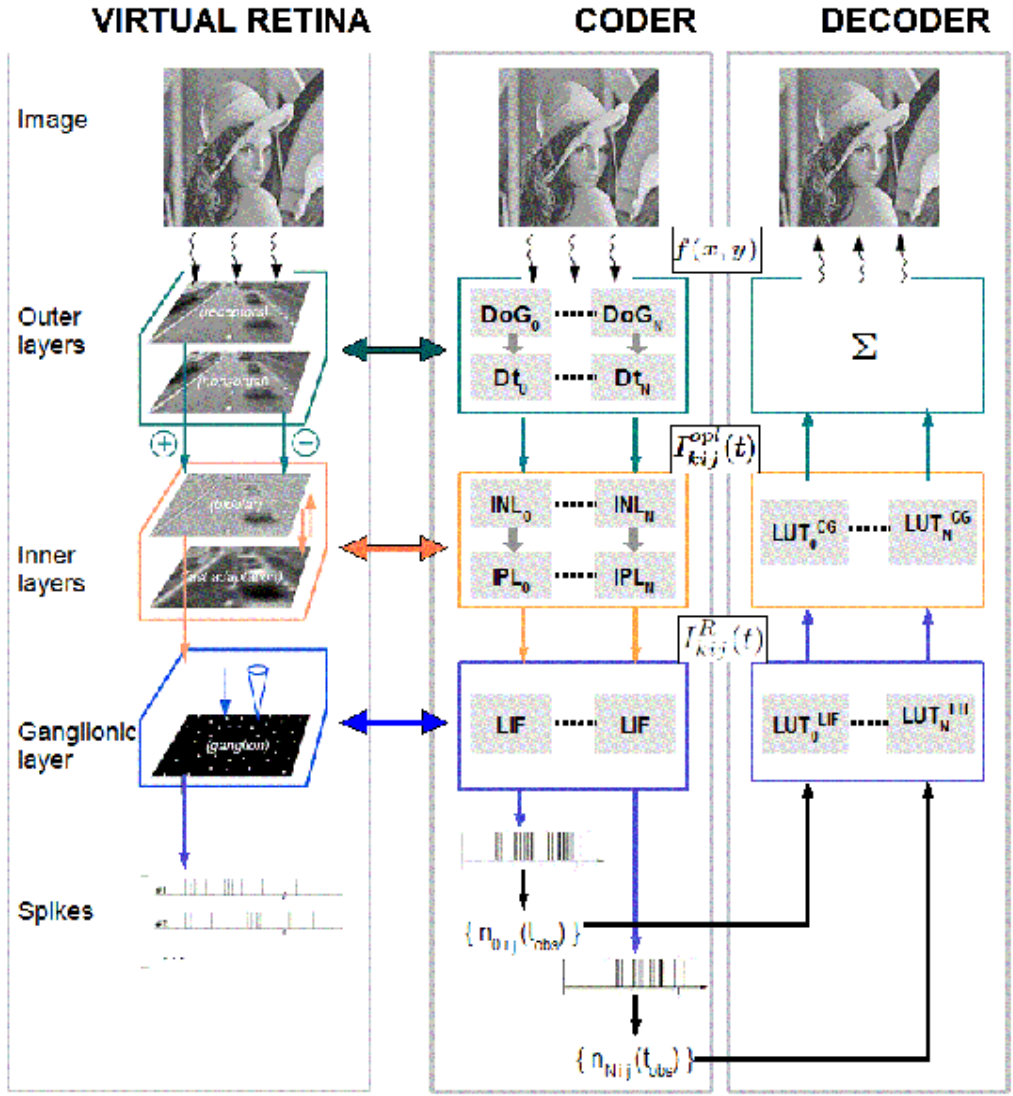

Figure 1: (a) Schematic view of the *Virtual Retina* model proposed by [36]. (b) and (c): Overview of our bio-inspired codec. Given an image, the static DoG-based multi-scale transform generates the subbands $\{F_k\}$. DoG filters are sorted from the lowest frequency-band filter $DoG_0$ to the highest one $DoG_{N-1}$. Each subband $F_k$ is delayed using a time-delay circuit $D_{t_k}$, with $t_k < t_{k+1}$. The time-delayed multi-scale output is then made available to the subsequent coder stages. The final output of the coder is a set of spike series, and the coding feature adopted will be the spike count $n_{kij}(t_{obs})$ recorded for each neuron indexed by $(kij)$ at a given time $t_{obs}$.

*3.1. The image transform: The outer layers of the retina*

In *Virtual Retina*, the outer layers were modelled by a non-separable spatio-temporal filtering. This processing produces responses corresponding to spatial or temporal variations of the signal because it models time-dependent interactions between two low-pass filters: this is termed center-surround differences. This stage has the property that it responds first to low spatial frequencies and later to higher frequencies. This *time-dependent frequency integration* was shown for *Virtual Retina* [37] and it was confirmed experimentally (see, e.g., [26]). This property is interesting as a large amount of the total signal energy is contained in the low frequencies subbands, whereas high frequencies bring further details. This idea already motivated bit allocation algorithms to concentrate the resources for a good recovery on lower frequencies.

However, it appears that inverting this non-separable spatio-temporal filtering is a complex problem [37, 38]. To overcome this difficulty, we propose to model differently this stage while keeping its essential features. To do so, we decomposed this process into two steps: The first one considers only center-surround differences in the spatial domain (through differences of Gaussians) which is justified by the fact that our coder here gets static images as input. The second step reproduces the time-dependent frequency integration by the introduction of time-delays.

*Center-surround differences in the spatial domain: The DoG model*

Neurophysiologic experiments have shown that, as for classical image coders, the retina encodes the stimulus representation in a transform domain. The retinal stimulus transform is performed in the cells of the outer layers, mainly in the outer plexiform layer (OPL). Quantitative studies such as [10, 24] have proven that the OPL cells processing can be approximated by a linear filtering. In particular, the authors in [10] proposed the largely adopted DoG filter which is a weighted difference of spatial Gaussians that is defined as follows:

$$DoG(x,y) = w_c G_{\sigma_c}(x,y) - w_s G_{\sigma_s}(x,y), \quad (1)$$

where $w_c$ and $w_s$ are the respective weights of the center and surround components of the receptive fields, and $\sigma_c$ and $\sigma_s$ are the standard deviations of the Gaussian kernels $G_{\sigma_c}$ and $G_{\sigma_s}$.

In terms of implementation, as in [28], the DoG cells can be arranged in a dyadic grid to sweep all the stimulus spectrum as schematized in Figure 2(a). Each layer $k$ in the grid, is tiled with $DoG_k$ cells having a scale $k$ and generating a transform subband $F_k$, where $\sigma_{s_{k+1}} = \frac{1}{2}\sigma_{s_k}$ and $\sigma_{c_{k+1}} = \frac{1}{2}\sigma_{c_k}$. So, in order to measure the degree of activation $\bar{I}^{opl}_{kij}$ of a given $DoG_k$ cell at the location $(i,j)$ with a scale $k$, we compute the convolution of the original image $f$ by the $DoG_k$ filter:

$$\bar{I}^{opl}_{kij} = \sum_{x,y=-\infty}^{\infty} DoG_k(i-x, j-y) f(x,y). \quad (2)$$

This transform generates a set of $(\frac{4}{3}N^2 - 1)$ coefficients for an $N^2$-sized image, as it works in the same fashion as a Laplacian pyramid [4]. An example of such a bio-inspired multi-scale decomposition is shown in Figure 2(b). Note here that we added to this bank of filters a Gaussian low-pass scaling function that represents the state of the OPL filters at the time origin. This yields a low-pass coefficient $\bar{I}^{opl}_{000}$ and enables the recovery of a low-pass residue at the reconstruction level [8, 19].

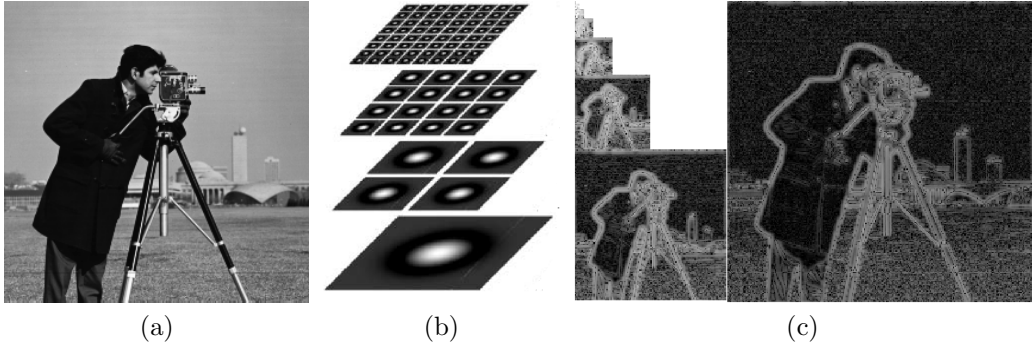

|  (a) | (b) | (c) |

Figure 2: (a) Input image cameraman. (b) Example of a dyadic grid of DoG's used for the image analysis (from [28]). (c) Example on image (a) of DoG coefficients generated by the retina model (the subbands are shown in the logarithmic scale)

*Integrating time dynamics through time-delay circuits*

Of course, the model described in (2) has no dynamical properties. In the actual retina, the surround $G_{\sigma_s}$ in (1) appears progressively across time driving the filter passband from low frequencies to higher ones. Our goal is to reproduce this phenomenon that we called time-dependent frequency integration. To do so, we added in the coding pathway of each subband $F_k$ a time-delay circuit $D_{t_k}$. The value of $t_k$ is specific to $F_k$ and is an increasing function of $k$. The $t_k$ delay causes the subband $F_k$ to be transmitted to the subsequent stages of the coder starting from the time $t_k$. The time-delayed activation coefficient $I^{opl}_{kij}(t)$ computed at the location $(i, j)$ for the scale $k$ at time $t$ is now defined as follows:

$$I^{opl}_{kij}(t) = \bar{I}^{opl}_{kij} \; \mathbb{1}_{\{t \geqslant t_k\}}(t), \tag{3}$$

where $\mathbb{1}_{\{t \geqslant t_k\}}$ is the indicator function such that, $\mathbb{1}_{\{t \geqslant t_k\}}(t) = 0$ if $t < t_k$ and 1 otherwise. While in our previous work [17] $t_k$ is increasing linearly as a function of $k$, we changed the law governing $t_k$ to an exponential one with a time constant denoted by $\tau^{opl}$. This change is intended to bring more biological plausibility to our new coder as the time behavior of the outer layers cells is exponential [10, 36]. Indeed, in the actual retina, the passband of the DoG cells runs through the low frequencies at a fast pace, then decelerates in an exponential fashion. So, the time-dependent frequency integration is not a linear phenomenon. The evolution of time delays $t_k$ with respect to the scale $k$, in the present work, is detailed in Figure 3.

### 3.2. The A/D converter: The inner and ganglionic layers of the retina

The retinal A/D converter is defined based on the processing occurring in the inner and ganglionic layers, namely a contrast gain control, a non-linear rectification and a discretization based on leaky integrate and fire neurons (LIF) [15]. A different treatment will be performed for each delayed subband, and this produces a natural bit allocation mechanism. Indeed, as each subband $F_k$ is presented at a different time $t_k$, it will be subject to a transform according to the state of our dynamic A/D converter at $t_k$.

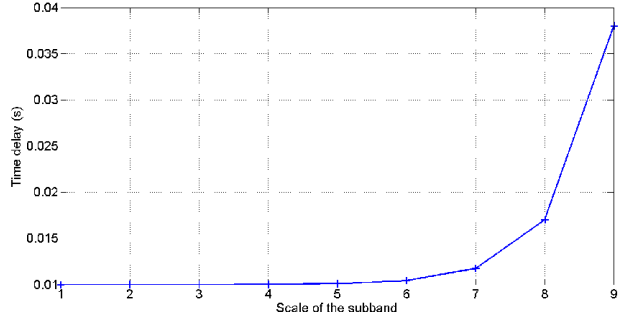

Figure 3: Time delays $D_{t_k}$ introduced in the coding process. The time-dependent frequency integration is reproduced by delaying the coding process start of the subband $F_k$ by $t_k$. The series $t_k$ is represented as a function of the scale $k$. The progression law is exponential with a time constant $\tau^{opl} = 65\,ms$.

*3.2.1. Contrast gain control*

The retina adjusts its operational range to match the input stimuli magnitude range. This is done by an operation called contrast gain control mainly performed in the bipolar cells. Indeed, real bipolar cells conductance is time varying, resulting in a phenomenon termed as *shunting inhibition*. This shunting avoids the system saturation by reducing high magnitudes.

In *Virtual Retina*, given the scalar magnitude $\bar{I}^{opl}_{kij}$ of the input step current $I^{opl}_{kij}(t)$, the contrast gain control is a non-linear operation on the potential of the bipolar cells. This potential varies according to both the time and the magnitude value $\bar{I}^{opl}_{kij}$; and will be denoted by $V^b_{kij}(t, \bar{I}^{opl}_{kij})$. This phenomenon is modelled, for a constant value of $\bar{I}^{opl}_{kij}$, by the following differential equation:

$$\begin{cases} c^b \dfrac{dV^b_{kij}(t, \bar{I}^{opl}_{kij})}{dt} + g^b(t) V^b_{kij}(t, \bar{I}^{opl}_{kij}) = I^{opl}_{kij}(t), & \text{for } t \geqslant 0, \\ g^b(t) = E_{\tau^b} \overset{t}{*} Q(V^b_{kij}(t, \bar{I}^{opl}_{kij})), \end{cases} \quad (4)$$

where $Q(V^b_{kij}) = g^b_0 + \lambda^b \left(V^b_{kij}(t)\right)^2$ and $E_{\tau^b} = \dfrac{1}{\tau^b} exp^{\frac{-t}{\tau^b}}$, for $t \geqslant 0$. Figure 4(a) shows the time behavior of $V^b_{kij}(t, \bar{I}^{opl}_{kij})$ for different magnitude values $\bar{I}^{opl}_{kij}$ of $I^{opl}_{kij}(t)$.

*3.2.2. Non-linear rectification*

In the next processing step, the potential $V^b_{kij}(t, \bar{I}^{opl}_{kij})$ is subject to a non-linear rectification yielding the so-called ganglionic current $I^g_{kij}(t, \bar{I}^{opl}_{kij})$. *Virtual Retina* models it, for a constant scalar value $\bar{I}^{opl}_{kij}$, by:

$$I^g_{kij}(t, \bar{I}^{opl}_{kij}) = N\left(T_{w^g, \tau^g}(t) * V^b_{kij}(t, \bar{I}^{opl}_{kij})\right), \quad \text{for } t \geqslant 0, \quad (5)$$

where $w^g$ and $\tau^g$ are constant scalar parameters, $T_{w^g, \tau^g}$ is the linear transient filter defined by $T_{w^g, \tau^g} = \delta_0(t) - w^g E_{\tau^g}(t)$, and $N$ is defined by:

$$N(v) = \begin{cases} \dfrac{i^g_0}{i^g_0 - \lambda^g(v - v^g_0)}, & \text{if } v < v^g_0 \\ i^g_0 + \lambda^g(v - v^g_0), & \text{if } v \geqslant v^g_0, \end{cases}$$

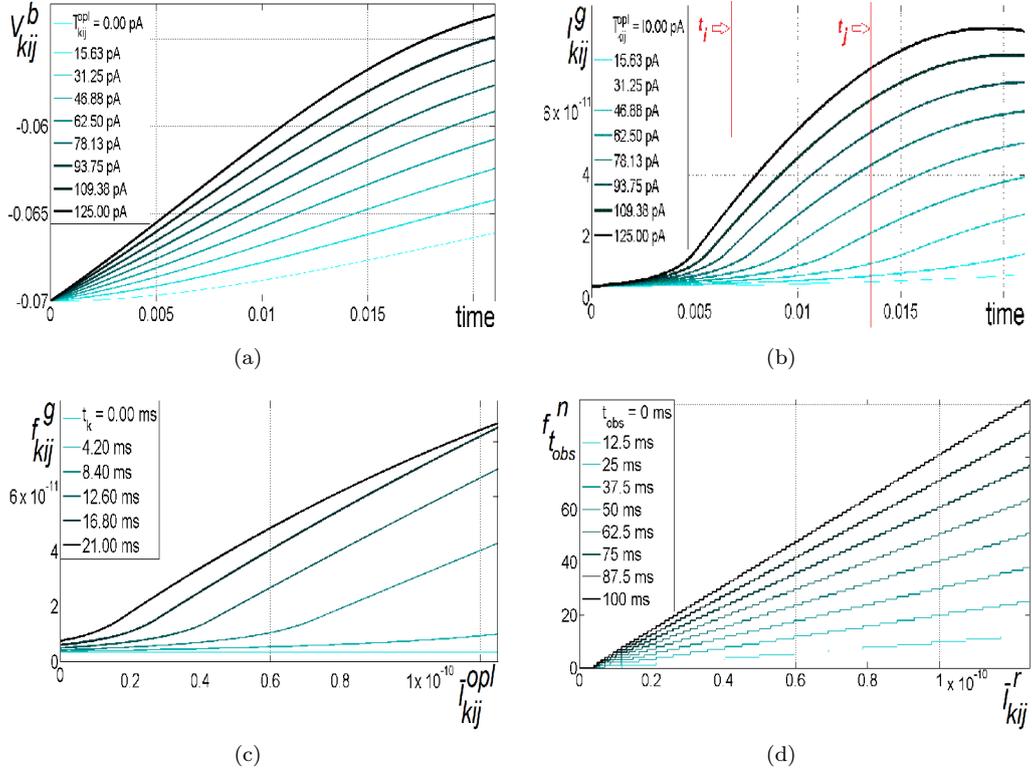

Figure 4: 4(a): $V^b_{kij}(t)$ as a function of time for different values of $\bar{I}^{opl}$; 4(b): $I^g_{kij}$ as a function of time for different values of $\bar{I}^{opl}$; 4(c): The functions $f^g_{t_k}$ that map $\bar{I}^{opl}_{kij}$ into $I^g_{kij}$ for different values of $t_k$; 4(d): The functions $f^n_{t_{obs}}$ that map $\bar{I}^r_{kij}$ into $n_{kij}$ for different values of $t_{obs}$

where $i^g_0$, $v^g_0$, and $\lambda^g$ are constant scalar parameters. Figure 4(b) shows the time behavior of $I^g_{kij}(t, \bar{I}^{opl}_{kij})$ for different values of $\bar{I}^{opl}_{kij}$.

As the currents $\bar{I}^{opl}_{kij}$ are delayed with times $\{t_k\}$, our goal is to catch the instantaneous behavior of the inner layers at these times $\{t_k\}$. This amounts to infer the transforms $I^g_{t_k}(\bar{I}^{opl}_{kij})$ that maps a given scalar magnitude $\bar{I}^{opl}_{kij}$ into a rectified current $\bar{I}^r_{kij}$ as the modelled inner layers would generate it at $t_k$. To do so, we start from the time-varying curves of $I^g_{kij}(t, \bar{I}^{opl}_{kij})$ in Figure 4(b) and we do a transversal cut at each time $t_k$. We show in Figure 4(c) the resulting maps $f^g_{t_k}$ such that $I^g_{kij}(t_k, \bar{I}^{opl}_{kij}) = f^g_{t_k}(\bar{I}^{opl}_{kij})$.

As for $I^{opl}_{kij}(t)$ (see Equation (3)), we introduce the time dimension using the indicator function $\mathbb{1}_{\{t \geq t_k\}}(t)$. The final output of this stage is the set of step functions $I^r_{kij}(t)$ defined by:

$$I^r_{kij}(t) = \bar{I}^r_{kij} \, \mathbb{1}_{\{t \geq t_k\}}(t), \text{ with } \bar{I}^r_{kij} = f^g_{t_k}(\bar{I}^{opl}_{kij}). \tag{6}$$

This non-linear rectification is analogous to a widely-used telecommunication technique: the companding [6]. Companders are used to make the quantization steps unequal after a linear gain control stage. Though, unlike $A-law$ or $\mu-law$ companders that amplify low magnitudes, the inner layers emphasize high magnitudes in the signal. Indeed, the bio-inspired compander, defined here, emphasizes high energy signals rather than high probability ones. This tendency is accentuated as the gain control gets higher across time. Besides, the inner layers stage have a time dependent behavior, whereas a usual gain controller/compander is static, and this makes our A/D converter go beyond the standards.

*3.2.3. Leaky integrate-and-fire quantization:*

The ganglionic layer is the deepest one tiling the retina: it transforms a continuous signal $I_{kij}^r(t)$ into discrete sets of spike trains. As in *Virtual Retina*, this stage is modelled by leaky integrate and fire neurons (LIF) which is a classical model. One LIF neuron is associated to every position in each subband $F_k$. The time-behavior of a LIF neuron is governed by the fluctuation of its voltage $V_{kij}(t)$. Whenever $V_{kij}(t)$ reaches a predefined $\delta$ threshold, a spike is emitted and the voltage goes back to a resting potential $V_R^0$. Between two spike emission times, $t_{kij}^{(l)}$ and $t_{kij}^{(l+1)}$, the potential evolves according to the following differential equation:

$$c^l \frac{dV_{kij}(t)}{dt} + g^l V_{kij}(t) = I_{kij}^r(t), \quad \text{for } t \in [t_{kij}^{(l)}, t_{kij}^{(l+1)}], \tag{7}$$

where $g^l$ is a constant conductance, and $c^l$ is a constant capacitance. In the literature, neurons activity is commonly characterized by the count of spikes emitted during an observation time bin $[0, t_{obs}]$, which we denote by $n_{kij}(t_{obs})$ [34]. Obviously, as $n_{kij}(t_{obs})$ encodes for the value of $I_{kij}^r(t)$, there is a loss of information as $n_{kij}(t_{obs})$ is an integer. The LIF is thus performing a quantization. If we observe the instantaneous behavior of the ganglionic layer at different times $t_{obs}$, we get a quasi-uniform scalar quantizer that refines in time. We can do this by a similar process to the one described in the previous paragraph. We show in Figure 4(d) the resulting maps $f_{t_{obs}}^n$ such that:

$$n_{kij}(t_{obs}) = f_{t_{obs}}^n(\bar{I}_{kij}^r). \tag{8}$$

Based on the set of spike counts $\{n_{kij}(t_{obs})\}$, measured at the output of our coder, we describe in the next section the decoding pathway to recover the initial image $f(x, y)$.

## 4. The decoding pathway

The decoding pathway is schematized in Figure 1(c). It consists in inverting, step by step, each coding stage described in Section 3. At a given time $t_{obs}$, the coding data is the set of $(\frac{4}{3}N^2 - 1)$ spike counts $n_{kij}(t_{obs})$, this section describes how we can recover an estimation $\tilde{f}_{t_{obs}}$ of the $N^2$-sized input image $f(x, y)$. Naturally, the recovered image $\tilde{f}_{t_{obs}}(x, y)$ depends on the time $t_{obs}$ which ensures time-scalability: the quality of the reconstruction improves as $t_{obs}$ increases. The ganglionic and inner layers are inverted using look-up tables constructed off-line and the image is finally recovered by a direct reverse transform of the outer layers processing.

*Recovering the input of the ganglionic layer:*

First, given a spike count $n_{kij}(t_{obs})$, we recover $\tilde{I}^r_{kij}(t_{obs})$, the estimation of $I^r_{kij}(t_{obs})$. To do so, we compute off-line the look-up table $n_{t_{obs}}(\bar{I}^r_{kij})$ that maps the set of current magnitude values $\bar{I}^r_{kij}$ into spike counts at a given observation time $t_{obs}$ (see Figure 4(d)). The reverse mapping is done by a simple interpolation in the reverse-look up table denoted $LUT^{LIF}_{t_{obs}}$. Here we draw the reader's attention to the fact that, as the input of the ganglionic layer is delayed, each coefficient of the subband $F_k$ is decoded according to the reverse map $LUT^{LIF}_{t_{obs}-t_k}$. Obviously, the recovered coefficients do not match exactly the original ones due to the quantization performed in the LIF's.

*Recovering the input of the inner layers:*

Second, given a rectified current value $\tilde{I}^r_{kij}(t_{obs})$, we recover $\tilde{I}^{opl}_{kij}(t_{obs})$, the estimation of $I^{opl}_{kij}(t_{obs})$. In the same way as for the preceding stage, we infer the reverse "inner layers mapping" through the pre-computed look up table $LUT^{CG}_{t_{obs}}$. The current intensities $\tilde{I}^{opl}_{kij}(t_{obs})$, corresponding to the retinal transform coefficients, are passed to the subsequent retinal transform decoder.

*Recovering the input stimulus:*

Finally, given the set of $(\frac{4}{3}N^2 - 1)$ coefficients $\{\tilde{I}^{opl}_{kij}(t_{obs})\}$, we recover $\tilde{f}_{t_{obs}}(x,y)$, the estimation of the original image stimulus $f(x,y)$. Though the dot product of every pair of $DoG$ filters is approximately equal to 0, the set of filters considered is not strictly orthonormal. We proved in [18] that there exists a dual set of vectors enabling an exact reconstruction. Hence, the reconstruction estimate $\tilde{f}$ of the original input $f$ can be obtained as follows:

$$\tilde{f}_{t_{obs}}(x,y) = \sum_{\{kij\}} \tilde{I}^{opl}_{kij}(t_{obs}) \widetilde{DoG}_k(i-x, j-y), \qquad (9)$$

where $\{kij\}$ is the set of possible scales and locations in the considered dyadic grid and $\widetilde{DoG}_k$ are the duals of the $DoG_k$ filters obtained as detailed in [18]. Equation (9) defines a progressive reconstruction depending on $t_{obs}$. This provides our code with an important feature: the scalability. Despite the fact that the input of our coder is a static image, we will be referring to this feature as time-scalability. Indeed, in our case different levels of rate and quality levels are achievable thanks to the observation time $t_{obs}$.

## 5. Results: Case of the bio-inspired and noiseless scalable image coder

We show examples of image reconstruction using our bio-inspired coder at different times[3]. Then, we study these results in terms of quality and bit-cost.
Quality is assessed by classical image quality criteria (PSNR and mean SSIM [31]). The cost is measured by the Shannon entropy $H(t_{obs})$ upon the population of $\{n_{kij}(t_{obs})\}$. The entropy computed in bits per pixel (*bpp*), for an $N^2$-sized image, is defined by:

$$H(t_{obs}) = \frac{1}{N^2} \sum_{k=0}^{K-1} 2^{2k} H\left(\left\{n_{s_k ij}(t_{obs}), (i,j) \in [\![0, 2^k - 1]\!]^2\right\}\right), \qquad (10)$$

where $K$ is the number of analyzing subbands. Figure 5 shows two examples of progressive reconstruction obtained with our new coder. Bit-rate/Quality are computed for each image in terms of the triplet (bit-rate in *bpp*/ PSNR quality in *dB*/ mean SSIM quality). Progressive reconstruction of cameraman in the left column yields: From top to bottom (0.006 *bpp*/ 16.02 *dB*/ 0.48), (0.077 *bpp*/ 18.34 *dB*/ 0.55), (0.23 *bpp*/ 21.20 *dB*/ 0.65), and (1.39 *bpp*/ 26.30 *dB*/ 0.84). Progressive reconstruction of baboon in the right column yields: From top to bottom (0.037 *bpp*/ 16.98 *dB*/ 0.18), (0.32 *bpp*/ 19.07 *dB*/ 0.35), (0.63 *bpp*/ 20.33 *dB*/ 0.49), and (2.24 *bpp*/ 27.37 *dB*/ 0.92).

---

[3] In all experiments, the model parameters are set to biologically realistic values: $g_0^b = 8\,10^{-10}\,S$, $\tau^b = 12\,10^{-3}\,s$, $\lambda^b = 9\,10^{-7}$, $c^b = 1.5\,10^{-10}\,F$, $v_0^g = 4\,10^{-3}\,V$, $i_0^g = 15\,10^{-12}\,A$, $w^g = 8\,10^{-1}$, $\tau^g = 16\,10^{-3}\,s$; $\lambda^g = 12\,10^{-9}\,S$, $\delta = 2\,10^{-3}\,V$, $g^L = 2\,10^{-9}\,S$, $V_R^0 = 0\,V$, $t_0 = 10\,10^{-3}\,s$, $t_{K-1} = 38\,10^{-3}\,s$, $\tau^{opl} = 65\,10^{-3}\,s$.

The new concept of *time scalability* is an interesting feature as it introduces time dynamics in the design of the coder. Figure 6 illustrates this concept. This is a consequence of the mimicking of the actual retina. We also notice that, as expected, low frequencies are transmitted first to get a first approximation of the image, then details are added progressively to draw its contours. The bit-cost of the coded image is slightly high. This can be explained by the fact that Shannon entropy is not the most relevant metric in our case as no context is taken into consideration, especially the temporal context. Indeed, one can easily predict the number of spikes at a given time $t$ knowing $n_{kij}(t - dt)$. Note also that no compression techniques, such that bit-plane coding, are yet employed. Our paper aims mainly at setting the basis of new bio-inspired coding designs.

For the reasons cited above, the performance of our coding scheme in terms of bit-cost have still to be improved to be competitive with the well established JPEG and JPEG 2000 standards. Thus we show no comparison in this paper. Though primary results are encouraging, noting that optimizing the bit-allocation mechanism and exploiting coding techniques as bit-plane coding [27] would improve considerably the bit-cost. Besides, the image as reconstructed with our bio-inspired coder shows no ringing and no block effect as in JPEG. Finally our codec enables scalability in an original fashion through the introduction of time dynamics within the coding mechanism.

Note also that differentiation in the processing of subbands, introduced through time-delays in the retinal transform, ensures an implicit bit-allocation mechanism. In particular the non-linearity in the inner layers stage amplifies singularities and contours, and these provide crucial information for the analysis of the image.

## 6. Introducing the noise in the coder: The non-subtractive dither hypothesis

In the preceding Sections 3 and 4, we presented the design of an image coder based on a bio-plausible model of the retina. We especially emphasized the deep retina layers analogy with A/D converters. Despite the fact that our coder takes into account several features of the actual retina as its time-dependent behavior, still it follows a deterministic law. Though, the actual neural code of the retina is clearly non-deterministic [9, 25]. Thus, in this section, we tackle the issue of the coding non-determinism in the retina. We make the proposal that the processing stages prior to the ganglionic layer yield a special type of noise: the dither noise. We then experience the perceptual impact of such a noise in our coder and give an original and plausible interpretation of its role in the stimuli coding process.

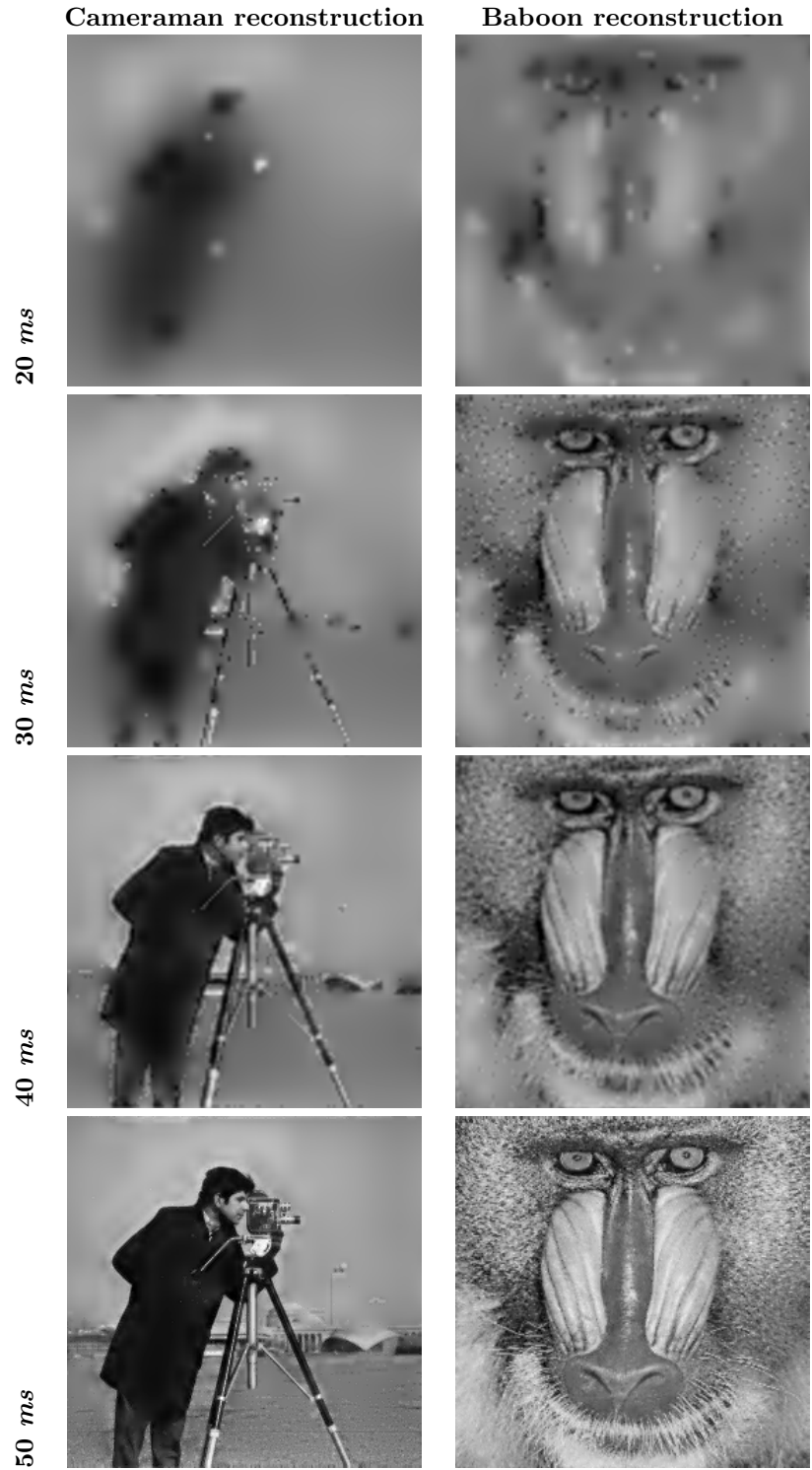

Figure 5: Progressive image reconstruction of cameraman and baboon using our new bio-inspired coder. The coded/decoded image is shown at: 20 ms, 30 ms, 40 ms, and 50 ms.

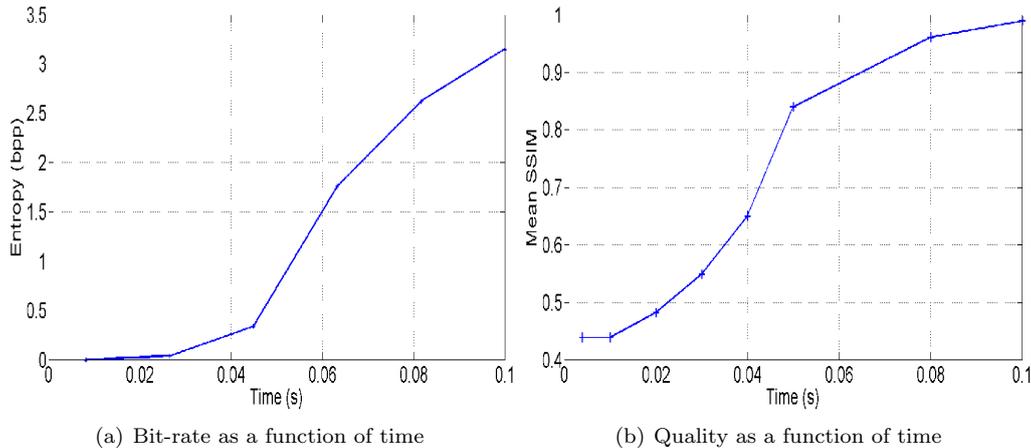

(a) Bit-rate as a function of time        (b) Quality as a function of time

Figure 6: Illustration of the concept of time scalability. the test image is cameraman. 6(a) shows the bit-rate variation of the encoded image as a function of the observation time $t_{obs}$. The bit-rate is measured by means of the entropy in bits per pixel (bpp). 6(b) shows the reconstruction quality variation as a function of the observation time $t_{obs}$. The quality is measured by means of the mean structural similarity index (mean SSIM). The only parameter that is tuned by the user is $t_{obs}$. Both quality and cost increase in accordance with $t_{obs}$. We talk about time-scalability.

*6.1. The non-deterministic nature of the neural code of the retina*

One major issue encountered by neuroscientists is the non-determinism of the retinal neural code. Indeed, given a single visual stimulus, spikes timings in the retina output are not exactly reproducible across trials. Yet, no clear evidence is established about the phenomena at the origin of this trial-to-trial variability. Several hypotheses were discussed in the literature and yielded two different points of view. The first hypothesis is that the precise timings of individual spikes convey a large amount of information [21, 20]. This hypothesis suggests that the stimulus coding process in the retina is deterministic and reports detailed information about the stimulus with a high temporal fidelity. In this case, each single spike timing makes sense. The second hypothesis is that only a few statistical quantities measured over the spike-based output convey the relevant information about the stimulus to the visual cortex [3]. For instance, since [1] it was widely assumed that the variable spike patterns corresponding to a single stimulus are random instantiations of a desired firing rate. In this case, the precise timing of each single spike may not be meaningful and thus spikes may carry some amount of noise. The spike based-output should then be averaged to reveal meaningful signals [9].

The role of spikes timings variability in the neural code of the retina is still an open issue and no clear evidence establishes whether this variability conveys precise information or random noise [25]. Here, we make the proposal that the non-determinism in the retinal processing prior to the ganglionic layer yields a dither noise [14, 16]. This noise, while corrupting the input of the ganglionic layer by a completely random signal, brings interesting features to the spike-based output of the retina. We will notice that the dithering process helps us to recognize the fine details of the stimulus earlier than with the noiseless coder. For this to be possible, the distribution of the noise that we introduce obeys specific constraints defined in [33]. Obviously, the dither noise hypothesis is one possible assumption among several others and we do not claim its biological exactness. Still, our present effort aims at bridging the differences between the different points of view reported above by exploring the hypothesis of a "retinal useful noise".

*6.2. Multiscale non-subtractive dithering*

We introduce in this section a multiscale dithering process that will be integrated in our bio-inspired image coder. Indeed, the coder that we designed has a multiscale architecture. So that the dither noise to be introduced must take into consideration the different scales of the retina model cells used for the image analysis.

We will assume that the processing stages of the retina that precede the ganglionic layer introduce a noise. As this noise is prior to the quantization done in the ganglionic layer, it is referred to as a dither noise. As, furthermore, this dither noise takes into consideration the multiscale architecture of the retina model, we will be talking about a multiscale dithering. The present work extends our previous efforts in [14, 16] to the multiscale case.

A few techniques referred to as multiscale dithering have been described in the literature. For example, in [30] the authors considered a hierarchical wavelet transform. The sibling subbands, ie. lying in the same level, are decorrelated by applying a series of rotations. The transform applied on the subbands is loosely referred to as dithering because it introduces a change on the wavelet coefficients prior to quantization. The resulting image is meant to reduce entropy while keeping the same perceptual quality. Another example is given in [12]. The authors used an image hierarchical quadtree representation and employ an error diffusion algorithm to get a binary halftone image. The distribution of binary pixels over the image space gives the impression of a multi-gray level image while using only two quantization levels. Although interesting, these state-of-the-art algorithms have one major drawback regarding the goals of our present work. Indeed, the techniques described rely on a totally deterministic algorithm. No random behavior is introduced during the coding process. Whereas in our case, we need to consider a coding process that may lead to different codes across trials for a single image. Besides the two algorithms are iterative and time consuming and this is contradicts the speed of processing in the retina.

In order to define the dither noise that corrupts the current $I^r_{kij}$ (cf. Equation (6)) at the input of the ganglionic layer, we reconsider the ganglion cell as a noisy leaky integrate and fire neuron (nLIF), that behaves according to the following equation:

$$c^l \frac{dV_{kij}(t)}{dt} + g^l V_{kij}(t) = I^r_{kij}(t) + \eta_{kij}, \quad \text{for } t \in [t^{(l)}_{kij}, t^{(l+1)}_{kij}], \tag{11}$$

The choice of the noise $\eta_{kij}$ distribution model to apply must obey two constraints: the biological plausibility and the mathematical constraints that provide our coder with interesting perceptual properties.

First, let us consider the biological plausibility constraint. Our aim is to mimic as closely as possible the actual retina behavior while modelling the multiscale dithering $\eta_{kij}$. So that, one must consider the nature of the dependency (if any) between the scale and the noise strength according to neurophysiologists observations. In this context, the authors in [13] stated that: *"The main difference between small and large cells is that the larger ones have lower peak sensitivity"*. This means that the large retina cells have a low reactivity to stimulus variations and thus are poorly affected by noise. On the contrary, small cells are extremely sensitive to stimulus variations and thus could be highly affected by noise. Our aim is to reproduce this phenomenon of noise strength variability as a function of retina cells scale. So that, we will corrupt the currents ($I^r_{kij}$) at the input of the retina ganglionic layer with noise coefficients $\eta_{kij}$, such that the dynamic range of this noise distribution depends on the cells scale $k$. The larger the subband $F_k$ cells are, the lower the noise dynamic range is. Thus, we will have to generate $K$ noise subbands, with an increasing dynamic range, to corrupt $K$ subbands of rectified currents $I^r_{kij}$.

Second, let us consider the mathematical constraint. Indeed we must consider the statistical properties that have to be verified by the added noise to provide our coder with interesting perceptual features. To this end, we refer to the results established in [33], and recall the following fundamental theorem of dither noise distribution for the case of a uniform scalar quantizer:

**Theorem 1.** *The choice of zero-mean dither probability distribution function (pdf) which renders the first and second moments of the total error independent of the input, such that the first moment is zero and the second is minimized, is unique and is a triangular pdf of 2 LSB peak-to-peak amplitude.*

Thus, we suppose that (i) $\eta_{kij}$ has a triangular probability distribution function with no loss of biological plausibility, and (ii) that the dynamic range of $\eta_{kij}$ is twice wider than the quantization step of the considered ganglion cell. Having these two conditions we verify the theorem. In this way, we identify the retinal noise $\eta_{kij}$ to a dither signal. As we do not subtract the dither signal in the decoding process, our coder is said to be a non-subtractive dithered system (NSD) [33, 32].

According to the discussion above we will consider that the noise $\eta_{kij}$ dynamic range (i) is an increasing function of the scale $k$ of the considered *DoG* retina cell, and (ii) is twice the width of the quantization step of the sussequent ganglion cell. Here we remind the reader that the ganglion cells are modelled, in our coder, by LIF neurons that are dynamic quantizers. Indeed the ganglionic layer evolves from a coarse to a fine quantizer. The quantization step of a LIF neuron will be denoted $Q^{lif}$. Obviously, $Q^{lif}$ is a decreasing function of the observation time $t_{obs}$ as shown in Figure 7. Furthermore, according to our original retina transform (cf. Section 3.1), the coding process of each subband $F_k$ is delayed in time by $t_k$. So that, the ganglion cells will have different levels of progression at a given time $t_{obs}$ depending on the subband scale $k$. We set the dithering parameters for an optimal observation time $t^*_{obs}$. So that, each subband will be corrupted by a noise subband having a triangular pdf which dynamic range $\Delta_k$ depends on the scale $k$ such that:

$$\begin{cases} Q^{lif}(t^*_{obs} - t_k) = Q^*_k \\ \Delta_k = 2\, Q^*_k \end{cases} \quad (12)$$

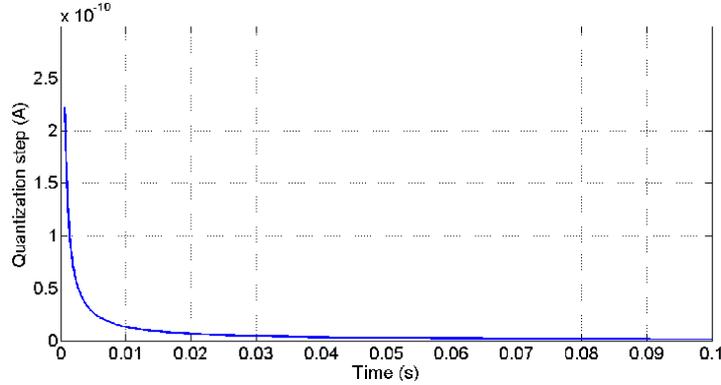

Figure 7: Estimation of the LIF neuron quantization step $Q^{lif}$ as a function of the observation time $t_{obs}$. The abscissa shows the observation times $t_{obs}$ between $0\,ms$ and $100\,ms$. The ordinate axis shows the mean quantization step $Q^{lif}$ estimated at a given $t_{obs}$ in Amperes.

An example of a multiscale dither noise thus defined is given in Figure 8. The test image is cameraman and the optimal observation time chosen is $t^*_{obs} = 52\,ms$. The rectified currents $I^r_{kij}$ in each subband of scale $k$ are subject to a dither noise $\eta_{kij}$ that has a triangular distribution with a dynamic range $\Delta_k$. We can notice that large cells in the low frequency subbands are poorly corrupted with noise while tight cells in the high frequency subbands are highly corrupted with noise. This is due to the fact that $\Delta_{k+1} > \Delta_k$, $\forall 0 \leqslant k < K - 2$. Interestingly, we remark that the time delays introduced in our model of the retinal transform allow us to implicitly satisfy the constraint of noise dynamic range $\Delta_k$ being an increasing function of the cells scale $k$.

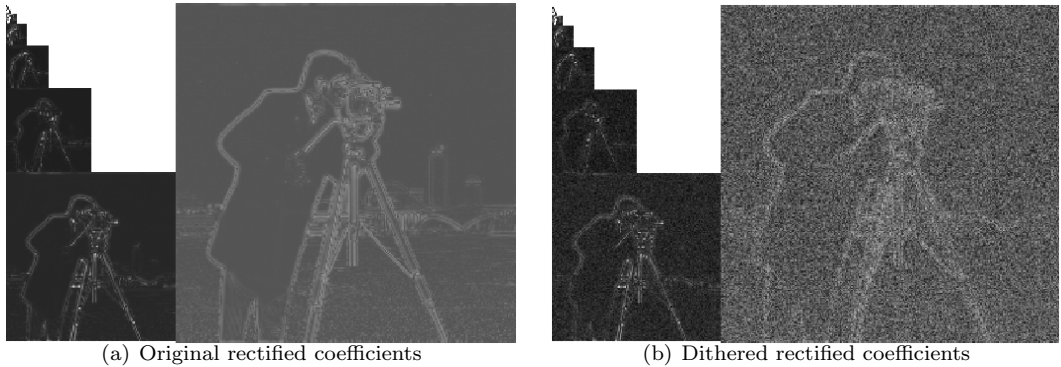

(a) Original rectified coefficients       (b) Dithered rectified coefficients

Figure 8: Example of dither noise introduced at the input of the ganglionic layer. The test image is cameraman. 8(a) shows the noiseless rectified coefficients $(I^r_{kij})$. 8(b) shows the rectified coefficients $(I^r_{kij} + \eta_{kij})$ with the dither noise $\eta_{kij}$ added. The noise parameters are set for the optimal observation time is $t^*_{obs} = 52ms$. The dither noise has a triangular distribution with a dynamic range $\Delta_k$ that depends on the subband $F_k$ considered. The larger the subband cells are, the lower the noise dynamic range is. The high frequencies are more corrupted with noise than the low frequencies.

Adding such a dither noise to the input of the ganglionic layer $I^r_{kij}$ induces interesting features. As specified in the theorem above, one important feature is the decorrelation between the reconstruction error at the output of the de-quantizer and the original signal at the input of the corresponding quantizer. The results of the theorem were demonstrated for uniform scalar quantizers. Whereas in our coder the ganglionic layer is not strictly a scalar quantizer but rather an approximation of it and, furthermore, the bio-inspired A/D converter that we designed is not uniform due to the preceding gain control and non-linear rectification stages. So that, we must verify the relevance of our approach. As the dithering process occurs in the DoG transform domain, we measure the error/input correlation in the transform domain. The error that we will denote by $\epsilon_{kij}$ is defined, in this case, as the difference between the output of the OPL layer $I^{opl}_{kij}$ and the estimation of it after decoding $\tilde{I}^{opl}_{kij}$, such that:

$$\epsilon_{kij} = I^{opl}_{kij} - \tilde{I}^{opl}_{kij} \qquad (13)$$

We can experimentally verify that, in fact, $\epsilon_{kij}$ and the input stimuli $I^{opl}_{kij}$ are decorrelated. This feature is clearly demonstrated when computing the cross correlation between $\epsilon_{kij}$ and $I^{opl}_{kij}$ as shown in Figures 9(a) and 9(b) for the test image cameraman and the highest frequency subband $F_{K-1}$. Comparable observations are made on the other subbands. Figure 9(a) shows the cross-correlation between $\epsilon_{kij}$ and $I^{opl}_{kij}$ measured for the noiseless case. The correlation is high especially when the spatial lag is small. 9(b) shows the same cross-correlation measures for the dithered case. We observe a very high decrease in the correlation even for the small spatial lags cases. Then, we can conclude that the signals $\epsilon_{kij}$ and $I^{opl}_{kij}$ are clearly decorrelated.

Another perceptually important feature that is induced by the dithering process is the error whitening. We verify also this feature in our case. As shown in Figures 9(c), the spectrum of the error obtained when using our coder with no addition of noise is non-uniform. This denotes strong geometric correlations in the error image which yields annoying artefacts. On the contrary, we notice in Figure 9(d) that the error spectrum is equally dispatched in the Fourier domain if we add a dither noise. Thus our new dithered scalable image coder gained interesting features through the integration of a dithering process.

The whitening and de-correlation features yield a greater reconstruction error in terms of mean squared error [33]. Though, the error whitening and decorrelation features acquired in the transform domain are perceptually important. Indeed, a strong correlation between the coding error and the original signal implies annoying artefacts. Besides the error whitening is important because all frequencies are affected by the same noise. The perceptual impact of dithering on the final image reconstruction $\tilde{f}_{t_{obs}}$ is shown in the next section.

## 7. Results: Case of the bio-inspired and dithered scalable image coder

We show in this section the perceptual impact of the dithering on the reconstructed images using our decoder. Our experiments demonstrate the ability of the dither noise

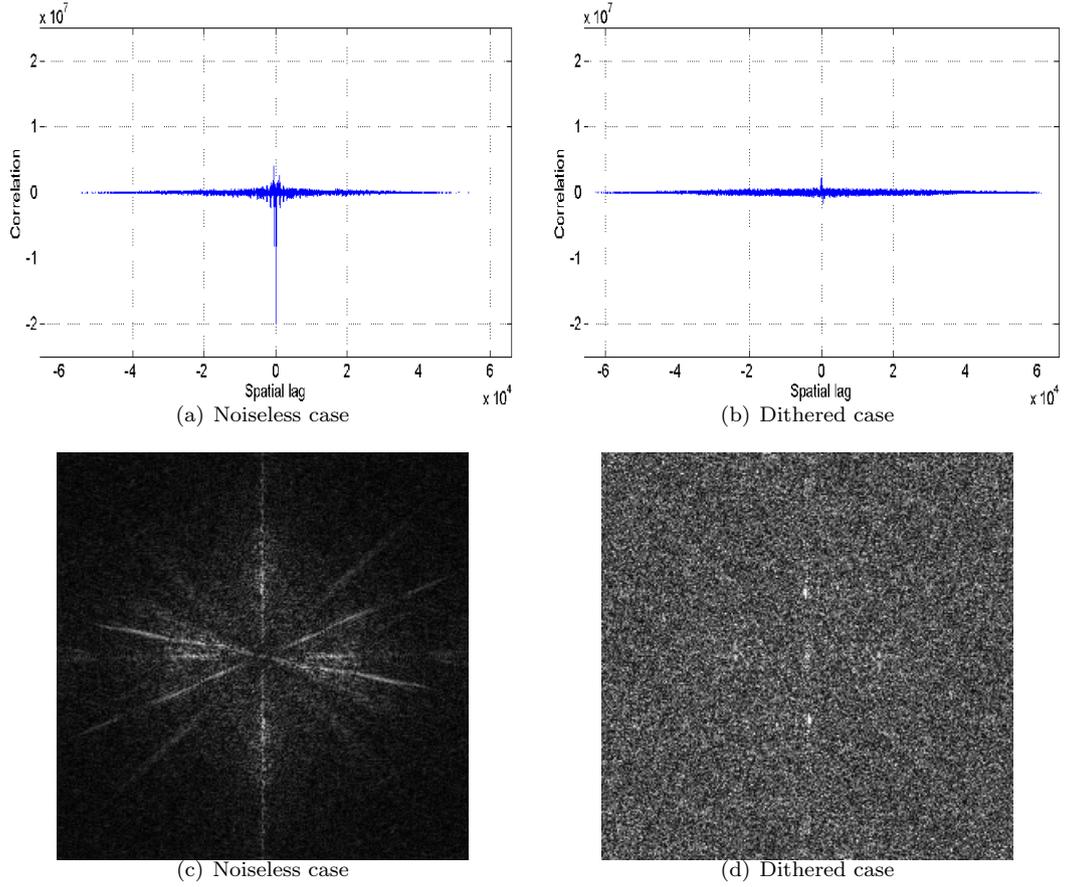

Figure 9: Error whitening and decorrelation in the DoG transform domain induced by the dither noise addition. The results are shown for the highest frequency subband $B_{K-1}$, but comparable observations are made on the other subbands. The dither noise is introduced at the input of the ganglionic layer. The dither noise parameters are set for the optimal observation time is $t^*_{obs} = 52ms$. 9(a) (respectively 9(b)) shows the cross-correlation between $\epsilon_{kij}$ and $I^{opl}_{kij}$ measured for the noiseless (resp. dithered) case. We observe a very high decrease in the correlation induced by the noise. The error is decorrelated from the input. 9(c) (respectively 9(d)) shows the amplitude spectrum of $\epsilon_{kij}$ computed for the noiseless (resp. dithered) case. We observe a wide spreading of the error spectrum in Fourier domain induced by the noise. The error is whitened.

to accelerate the recognition of the image details and singularities during the decoding process.

A first example is given in Figure 10. The left column shows the evolution of the reconstruction $\tilde{f}_{t_{obs}}$ with increasing times $t_{obs}$, in the case of noiseless coding. The right column shows the evolution of the reconstruction $\tilde{f}_{t_{obs}}$ with increasing times $t_{obs}$, in the case of addition of a dither noise to the input of the ganglionic layer. The central column shows a filtered version of cameraman. Cameraman is sharpened to enhance the image details. The comparison between the noiseless case reconstruction (on the left) and the dithered reconstruction (on the right) demonstrates perceptual importance of noise in the image coding process in the retina. With the addition of noise, details of cameraman are well rendered "before date". For example, the hand of cameraman and the tower in the background appear since $t_{obs} = 44ms$ for the dithered case while still invisible in the noiseless case at the same observation time. We can also notice that the horizontal stripes in the background, the grass details, the pant folds, and the hand are well rendered since $t_{obs} = 48ms$. On the contrary these details are still invisible or highly blurry in the noiseless case at the same observation time. Finally, at the optimal observation $t_{obs} = t^*_{obs} = 52ms$ all the fine details of the image, including the coat and the background details, are clearly distinguished in the dithered case while still blurry or invisible in the noiseless case.

A second example is given in Figure 11 for the baboon test image. This image is rich of details and singularities and thus particularly challenging. Though, our dithered coder still renders the image details "before date in this case" (with another adequate parametrization for the dither noise). As for the preceding example, the left image shows the reconstruction $\tilde{f}_{t_{obs}}$ in the case of noiseless coding. The right image shows the reconstruction $\tilde{f}_{t_{obs}}$ in the case of addition of a dither noise to the input of the ganglionic layer. The central image is a sharpened version of baboon. The observation time shown in this figure is also the optimal observation $t_{obs} = t^*_{obs} = 44\,ms$. The comparison between the noiseless case reconstruction (on the left) and the dithered reconstruction (on the right) confirms the observations made in the first example. The dither noise helps the recognition of fine details "before date". While in the noiseless case face and bear details of baboon are still blurry, these details are well rendered in the dithered reconstruction case.

On one hand, the integration of a dither noise in the coding process yield a greater reconstruction error in terms of mean squared error [33]. Besides, as the dither noise is a disordered signal, it also increases the entropy of the image code. On the other hand, the error whitening and de-correlation features acquired by our system are perceptually important. This is a crucial point because our current results may prove that the retina conveys a code that is optimized for the tasks to be performed by the visual cortex as categorization. While the rate/distortion trade-off remains an important goal for a coding scheme it may not be the central performance criterion for the retina.

## 8. Conclusion

The work that we presented brings two main contributions. As a first step, we proposed a bio-inspired codec for static images with a deterministic behavior. The image coder is based on two stages. The first stage is the image transform as performed by the outer layers of the retina. In order to integrate time dynamics, we added to this transform time delays that are subband specific so that, each subband is processed differently. The second stage is a succession of two dynamic processing steps mimicking the deep retina layers behavior. These latter perform an A/D conversion and generate a spike-based, invertible, retinal code for the input image in an original fashion.

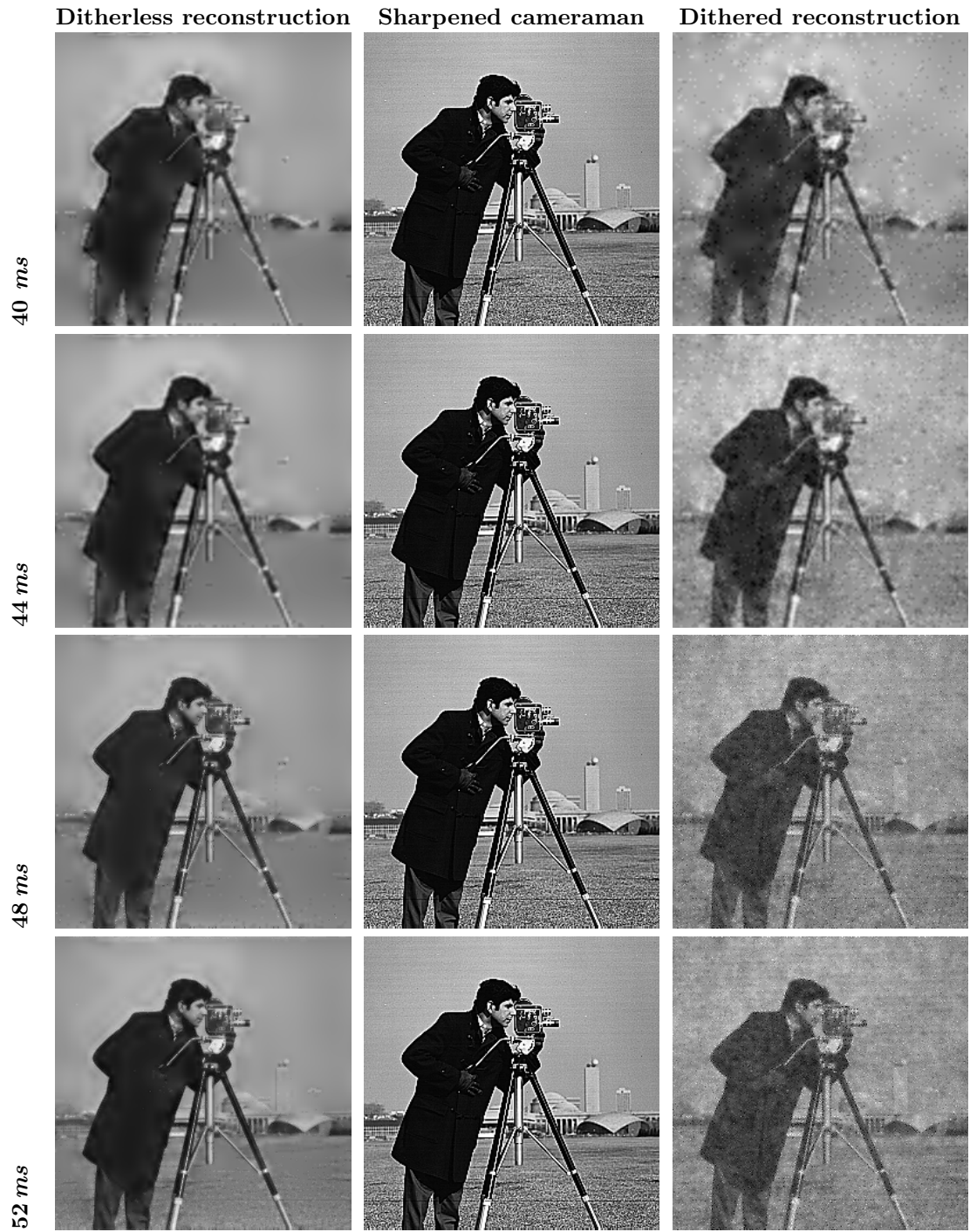

Figure 10: Perceptual impact of the multiscale dithering on the reconstruction of cameraman. The observation times $t_{obs}$ are shown on the left. From top to bottom, $t_{obs}$ take successively the values of: $40\,ms$, $44\,ms$, $48\,ms$, and $52\,ms$. The observation time shown in this figure is also the optimal observation $t_{obs} = t_{obs}^* = 44\,ms$.

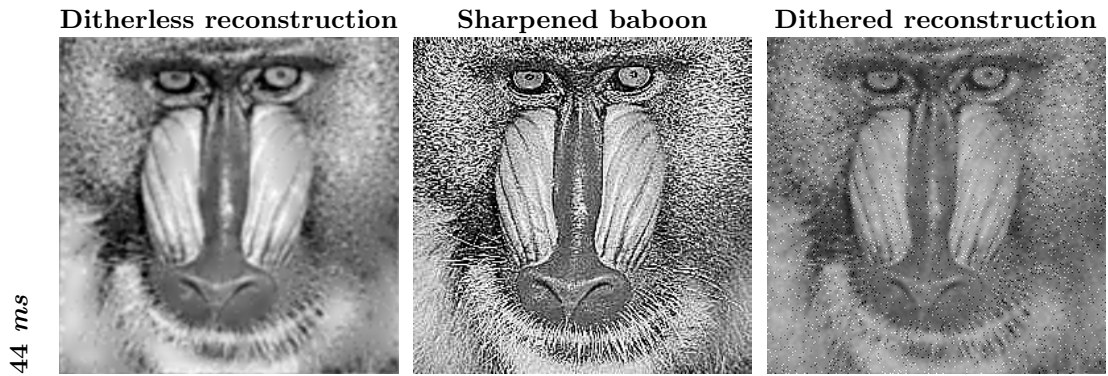

Figure 11: Perceptual impact of the multiscale dithering on the reconstruction of baboon. The observation time shown in this figure is also the optimal observation $t_{obs} = t^*_{obs} = 44\,ms$.

In a second step, we investigated the issue of non-determinism in the retina neural code. We proposed to model the retinal noise by a multiscale dither signal with specific statistical properties. The dithering process that we proposed whitens the reconstruction error and decorrelates it from the input stimuli. Besides, from a perceptual point of view, our coder allows an earlier recognition of the image details and singularities during the decoding process.

In conclusion, our coding scheme offers interesting features such as (i) time-scalability, as the choice of the observation time of our codec enables different reconstruction qualities, and (ii) bit-allocation, as each subband of the image transform is separately mapped according to the corresponding state of the inner layers. In addition, when integrating a dithering process our coder gained interesting perceptual features. These features, if the dithering hypothesis is confirmed, help the visual cortex recognize the fine details of the image. This latter point is interesting because it may prove that the retina conveys a code that is optimized for the tasks to be performed by the visual cortex. Interestingly, our dithering hypothesis found an echo recently in the computational neurosciences community [29]. We are convinced that further neurophysiologic investigations may also confirm the relevance of dithering in the retinal processing.

In terms of rate/distortion, the results accomplished by our coding scheme are encouraging. Though the rate/distortion performance is not the primary goal of this work, our coder could still be improved to be competitive with the well established JPEG and JPEG 2000 standards. Optimizing techniques as bit-plane coding are to be investigated.

This work is at the crossroads of diverse hot topics in the fields of neurosciences, brain-machine interfaces, and signal processing and tries to bridge the gap between these different domains towards the conception of new biologically inspired coders.